\documentclass[submission,copyright,creativecommons]{eptcs}
\providecommand{\event}{FROM 2023} 

\usepackage{iftex}

\ifpdf
  \usepackage[T1]{fontenc}        
  \usepackage{amsmath,amsthm,amssymb,graphicx,hyperref}
  \usepackage[english]{babel}
  \usepackage{paralist}
  \usepackage{graphicx}
  \usepackage{amsmath,amssymb}
  \usepackage{xcolor}
    \usepackage{paralist}
    \usepackage{caption}
    \usepackage{subcaption}
    \usepackage{booktabs}
    \usepackage{hyperref}
    \usepackage{multirow}
  \usepackage{xcolor}
  \theoremstyle{definition}
\newtheorem{defn}{Definition}[section]
\else
  \usepackage{breakurl}           
\fi

\title{Benchmarking Local Robustness of High-Accuracy Binary Neural Networks for Enhanced Traffic Sign Recognition}
\author{
Andreea Postovan \qquad\qquad M\u{a}d\u{a}lina Era\c{s}cu
\institute{Faculty of Mathematics and Informatics, West University of Timisoara \\ 
           4 blvd. V. Parvan, 300223, Romania}
\email{\{andreea.postovan99, madalina.erascu\}@e-uvt.ro}
}
\def\titlerunning{Benchmarking Local Robustness of High-Accuracy BNNs for Enhanced Traffic Sign Recognition}
\def\authorrunning{Postovan \& Era\c{s}cu}
\begin{document}
\maketitle

\begin{abstract}
Traffic signs play a critical role in road safety and traffic management for autonomous driving systems. Accurate traffic sign classification is essential but challenging due to real-world complexities like adversarial examples and occlusions. To address these issues, binary neural networks offer promise in constructing classifiers suitable for resource-constrained devices.

In our previous work, we proposed high-accuracy BNN models for traffic sign recognition, focusing on compact size for limited computation and energy resources. To evaluate their local robustness, this paper introduces a set of benchmark problems featuring layers that challenge state-of-the-art verification tools. These layers include binarized convolutions, max pooling, batch normalization, fully connected. The difficulty of the verification problem is given by the high number of network parameters (905k - 1.7 M), of the input dimension (2.7k-12k), and of the number of regions (43) as well by the fact that the neural networks are not sparse.

The proposed BNN models and local robustness properties can be checked at \url{https://github.com/ChristopherBrix/vnncomp2023_benchmarks/tree/main/benchmarks/traffic_signs_recognition}. 

The results of the 4th International Verification of Neural Networks Competition (VNN-COMP'23) revealed the fact that 4, out of 7, solvers can handle many of our benchmarks randomly selected (minimum is 6, maximum is 36, out of 45). Surprisingly, tools output also wrong results or missing counterexample (ranging from 1 to 4).
Currently, our focus lies in exploring the possibility of achieving a greater count of solved instances by extending the allotted time (previously set at 8 minutes). Furthermore, we are intrigued by the reasons behind the erroneous outcomes provided by the tools for certain benchmarks.


\end{abstract}
\section{Introduction}
Traffic signs play a crucial role in ensuring road safety and managing traffic flow, both in urban and highway driving. For autonomous driving systems, the accurate recognition and classification of traffic signs, known as \emph{traffic sign classification (recognition)}, are essential components. This process involves two main tasks: firstly, isolating the traffic sign within a bounding box, and secondly, classifying the sign into a specific traffic category. The focus of this work lies on the latter task.

Creating a robust traffic sign classifier is challenging due to the complexity of real-world traffic scenes. Common issues faced by classifiers include a lack of \emph{robustness} against \emph{adversarial examples}~\cite{szegedy2013intriguing} and occlusions~\cite{zhang2020lightweight}. \emph{Adversarial examples} are inputs that cause classifiers to produce erroneous outputs, and \emph{occlusions} occur naturally due to various factors like weather conditions, lighting, and aging, which make traffic scenes unique and diverse.

To address the lack of robustness, one approach is to formally verify that the trained classifier can handle both adversarial and occluded examples. Binary neural networks (BNNs) have shown promise in constructing traffic sign classifiers, even in devices with limited computational resources and energy constraints, often encountered in autonomous driving systems. BNNs are neural networks (NNs) with binarized weights and/or activations constrained to $\pm 1$, reducing model size and simplifying image recognition tasks.

The long-term goal of this work is to provide formal guarantees of specific properties, like robustness, that hold for a trained classifier. This objective leads to the formulation of the \emph{verification problem}: given a trained model and a property to be verified, does the model satisfy that property? The verification problem is translated into a constrained satisfaction problem, and existing verification tools can be employed to solve it. However, due to its \textsc{NP}-complete nature \cite{katz2022reluplex}, this problem is experimentally challenging for state-of-the-art tools.

In our previous work \cite{postovan2023architecturing}, we proposed high-accuracy BNN models explicitly for traffic sign recognition, with a thorough exploration of accuracy, model size, and parameter variations for the produced architectures. The focus was on BNNs with high accuracy and compact model size, making them suitable for devices with limited computation and energy resources, while also reducing the number of parameters to facilitate the verification task. The German Traffic Sign Recognition Benchmark (GTSRB)~\cite{GTSRB} was used for training, and testing involved similar images from GTSRB, as well as Belgian~\cite{BelgianTrafficSignDatabase} and Chinese~\cite{ChineseTrafficSignDatabase} datasets. This paper builds upon the models with the best accuracy from the previous study \cite{postovan2023architecturing} and presents a set of benchmark problems to verify local robustness properties of these models.

The novelty of the proposed benchmarks lies in the fact that traffic signs recognition is done using binarized neural networks. To the best of our knowledge this was not done before \cite{ciregan2012multi,stallkamp2012man}. Compared to existing benchmarks. The types of layers used determine a complex verification problem and include \emph{binarized convolution layers} to capture advanced features from the image dataset, \emph{max pooling layers} for model size reduction while retaining relevant features, \emph{batch normalization layers} for scaling, and \emph{fully connected (dense) layers}. The difficulty of the verification problem is given by the high number of network parameters (905k - 1.7 M), of the input dimension (2.7k-12k), and of the number of regions (42) as well by the fact that the neural networks are not sparse.
Discussions with organizers and competitors in the Verification of Neural Network Competition (VNN-COMP)\footnote{https://github.com/stanleybak/vnncomp2023/issues/2} revealed that no tool competing in 2022 could handle the proposed benchmark. Additionally, in VNN-COMP 2023~\cite{BenchmarksVNN-COMP-23}, the benchmark was considered fairly complex by the main developer of the winning solver $\alpha, \beta$-CROWN\footnote{https://github.com/Verified-Intelligence/alpha-beta-CROWN}. 

We publicly released our bechmark in May 2023. In the VNN-COMP 2023, which took place in July 2023, our benchmark was used in scoring, being nominated by at least 2 competing tools. 4, out of 7, tools were able to find an answer for the randomly generated instances. Most instances were solved by $\alpha, \beta$-CROWN (39 out of 45) but it received penalties for 3 results due to either incorrect answer or missing counterexample. Most correct answers were given by Marabou\footnote{https://github.com/NeuralNetworkVerification/Marabou} (18) with only 1 incorrect answer.

Currently, we are investigating the reasons why the tools were not able to solve all instances and why incorrect answers were given. Additionally, more tests will be performed on randomly generated answers and we will examine the particularities of the input images and of the trained networks which can not be handled by solvers due to timeout or incorrect answer.

\smallskip

The rest of the paper is organized as follows. In Section~\ref{sec:RelWork} we present related work focusing on comparing the proposed benchmark with others competing in VNN-COMP. Section~\ref{sec:TheoreticalBkg} briefly describes deep neural networks, binarized neural networks and formulates the robustness property. In Section~\ref{sec:anatomy} we describe the anatomy of the trained neural networks whose local robustness is checked. In Section \ref{sec:modelRepresentation} we introduce the verification problem and its canonical representation (VNN-LIB and ONNX formats). Section~\ref{sec:Bench&ExpResults} presents the methodology for benchmarks generation and the results of the VNN-COMP 2023.
\section{Related Work}\label{sec:RelWork}
There exist many approaches for the verification of neural networks, see ~\cite{Zhang_tutorial} for a survey, however few are tackling the verification of binarized neural networks.

Verifying properties using boolean encoding~\cite{ narodytska2018verifying} is an alternative approach to validate characteristics of a specific category of neural networks, known as binarized neural networks. These networks possess binary weights and activations. The proposed technique involves reducing the verification problem from a mixed integer linear programming problem to a Boolean satisfiability. By encoding the problem in Boolean logic, they exploit the capabilities of modern SAT solvers, combined with a counterexample-guided search method, to verify various properties of these networks. A primary focus of their research is assessing the networks' resilience against adversarial perturbations. The experimental outcomes demonstrate the scalability of this approach when applied to medium-sized deep neural networks employed in image classification tasks. However their neural networks do not have convolution layers and can handle only a simple dataset like \textsc{MNIST} where images are black and white and there are just $10$ classes to classify. Also, no tool implementing the approach was realeased to be tested.

Paper~\cite{amir2021smt} focuses on verification of binarized neural network, extended the Marabou~\cite{katz2022reluplex} tool to support \emph{Sign Constrains} and verified a network that uses both binarized and non-binarized layers. For testing they used Fashion-\textsc{MNIST} dataset which was trained using \textsc{XNOR-Net} architecture and obtained the accuracy of only 70.97\%. This extension could not be used in our case due to the fact that we have binarized convolution layers which the tool can not handle.

In the verification of neural networks competition (VNN-COMP), in 2022, there are various benchmarks subject to verification \cite{BenchmarksVNN-COMP-22}, however, there is none involving traffic signs. To the best of our knowledge there is only one paper which deals with traffic signs datasets \cite{guo2023occrob} that is GTSRB. However, they considered only subsets of the dataset and their trained models consist of only fully connected (FC) layers with ReLU activation functions, not convolutions, ranging from 70 to 1300 neurons. Furthermore they do not mention the accuracy of their trained models to be able to compare it with ours. Moreover, the benchmarks from VNN-COMP~2022~\cite{müller2023international}  used for image classification tasks have are in Table~\ref{tab:bench-VNN-COMP-2022}. As one could observe, no benchmarks use binarized convolutions and batch normalization layers. Discussions with competition organizers revealed the fact that no tool from 2022 competition could handle our benchmark\footnote{See https://github.com/stanleybak/vnncomp2023/issues/2 intervention from user stanleybak on May 17, 2023}.
\begin{table}[h]
\centering
\scriptsize
\caption{Benchmarks proposed in the VNN-COMP 2022 for image classification tasks}
\label{tab:bench-VNN-COMP-2022}
\begin{tabular}{ccccc}
\textbf{Category}              & \textbf{Benchmark} & \textbf{Network Types} & \textbf{\#Neurons} & \textbf{Input Dimension} \\ \hline
\multirow{5}{*}{CNN \& ResNet} & Cifar Bias Field   & Conv. + ReLU           & 45k                & 16                       \\
                               & Large ResNets      & ResNet (Conv. + ReLU)  & 55k - 286k         & 3.1k - 12k               \\
                               & Oval21             & Conv. + ReLU           & 3.1k - 6.2k        & 3.1k                     \\
                               & SRI ResNet A/B     & ResNet (Conv. + ReLU)  & 11k                & 3.1k                     \\
                               & VGGNet16           & Conv. + ReLU + MaxPool & 13.6M              & 1 - 95k                  \\ \hline
Fully-Connected               & MNIST FC           & FC. + ReLU             & 512 - 1.5k         & 784                   
\end{tabular}
\end{table}

The report of this year neural networks verification competition (VNN-COMP 2023) is in the draft version, but we present here the differences between our benchmark and the others. Table \ref{tab:VNN-COMP-2023-benchmarks} taken from the draft report presents all the scored benchmarks, i.e. benchmarks which were nominated by at least 2 competing tools and are used in their ranking. The column Network Type presents the types of layers of the trained neural network, the column \# of Params represent the number of parameters of the trained neural network, the column Input Dimension represents the dimension of the input (for example, for an image with dimension 30x30 pixels and RGB channel the dimension is 30x30x3 which means that the verification problem contains 30x30x3 variables), the Sparsity column represents the degree of sparsity of the trained neural network and, finally, the column \# of Regions represents the number of regions determined by the verification problem (for example, for our German Traffic Sign Recognition Benchmark there are 43 traffic signs classes). Our proposed benchmark, Traffic Signs Recognition, is more complex as the others as it involves cumulatively a high number of parameters, input dimension, number of regious and no sparsity.

\begin{table}[h]
\centering
\scriptsize
\caption{Benchmarks proposed in the VNN-COMP 2023}
\label{tab:VNN-COMP-2023-benchmarks}
\begin{tabular}{cccccc}
\textbf{Name}             & \textbf{Network Type}                                                                                   & \textbf{\# of Params} & \textbf{\begin{tabular}[c]{@{}c@{}}Input \\ Dimension\end{tabular}} & \textbf{Sparsity} & \textbf{\# of Regions} \\ \hline
nn4sys                    & Conv, FC, Residual + ReLU, Sigmoid                                                                      & 33k - 37M             & 1-308                                                               & 0-66\%   & 1 - 11k                \\ \hline
VGGNet16                  & Conv + ReLU + MaxPool                                                                                   & 138M                  & 150k                                                                & 0-99\%   & 1                      \\ \hline
Collins Rul CNN           & Conv + ReLU, Dropout                                                                                    & 60k - 262k            & 400-800                                                             & 50-99\%  & 2                      \\ \hline
TLL Verify Bench          & FC + ReLU                                                                                               & 17k - 67M             & 2                                                                   & 0\%      & 1                      \\ \hline
Acas XU                   & FC + ReLU                                                                                               & 13k                   & 5                                                                   & 0-20\%   & 1-4                    \\ \hline
cGAN                      & \begin{tabular}[c]{@{}c@{}}FC, Conv, ConvTranspose, \\ Residual + ReLU, BatchNorm, AvgPool\end{tabular} & 500k-68M              & 5                                                                   & 0-40\%   & 2                      \\ \hline
Dist Shift                & FC + ReLU, Sigmoid                                                                                      & 342k-855k             & 792                                                                 & 98.9\%   & 1                      \\ \hline
ml4acopf                  & FC, Residual + ReLU, Sigmoid                                                                            & 4k-680k               & 22-402                                                              & 0-7\%    & 1-600                  \\ \hline
\textcolor{blue}{Traffic Signs Recogn} & \textcolor{blue}{Conv+Sign+MaxPool+BatchNorm, FC,}                                                                        & \textcolor{blue}{905k-1.7M}             & \textcolor{blue}{2.7k-12k}                                                            & \textcolor{blue}{0\%}      & \textcolor{blue}{43}                     \\ \hline
ViT                       & Conv, FC, Residual + ReLU, Softmax, BatchNorm                                                           & 68k-76k               & 3072                                                                & 0\%      & 9                      \\ \hline
\end{tabular}
\end{table}
\section{Theoretical Background}\label{sec:TheoreticalBkg}

\subsection{Deep Neural Networks}\label{sec:NNs}
\emph{Neural networks}, inspired by the human brain, are computational models composed of interconnected nodes called artificial neurons. These networks have gained attention for their ability to learn and perform complex tasks. The nodes compute outputs using \emph{activation functions}, and synaptic \emph{weights} determine the strength of connections between nodes. Training is achieved through optimization algorithms, such as \emph{backpropagation}, which adjust the weights iteratively to minimize the network's error.

A \emph{deep neural network (DNN)}~\cite{amir2021smt} can be conceptualized as a directed graph, where the nodes, also known as neurons, are organized in \emph{layers}. The input layer is responsible for receiving initial values, such as pixel intensities in the case of image inputs, while the output layer generates the final predictions or results. Hidden layers, positioned between the input and output layers, play a crucial role in extracting and transforming information.
\begin{figure}[h!]
\centering
\includegraphics[width=0.45\textwidth]{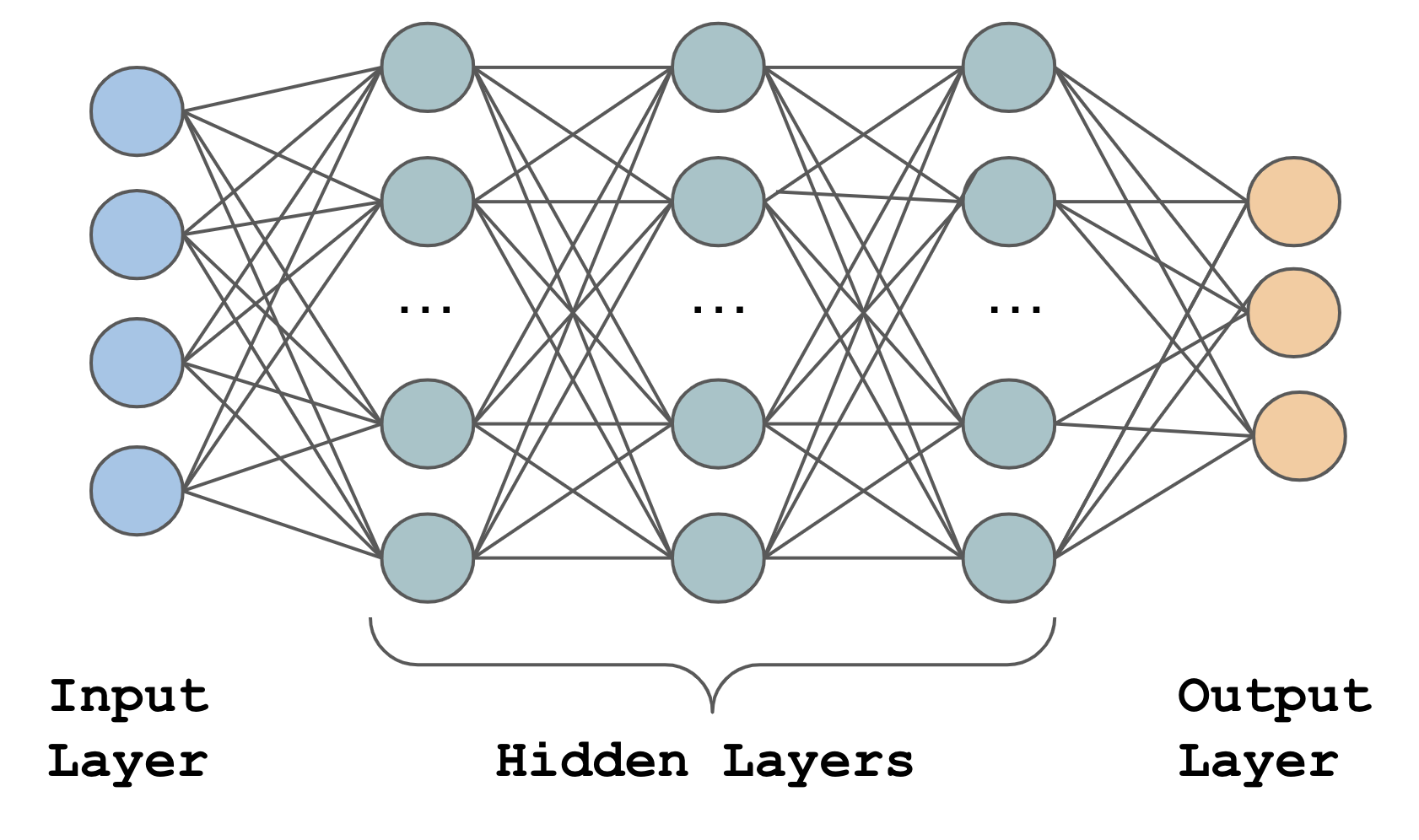}
\caption{A fully connected DNN with 4 input nodes, 3 output nodes and 3 hidden layers}
\label{fig:dnn}
\end{figure}
During the evaluation or inference process, the input values propagate through the network, layer by layer, using connections between neurons. Each neuron applies a specific mathematical operation to the inputs it receives, followed by the \emph{activation function} that introduces \emph{non-linearity} to the network. The activation function determines the neuron's output based on the weighted sum of its inputs and an optional bias term.

Different layer types are employed in neural networks to compute the values of neurons based on the preceding layer's neuron values. Those relevant for our work are introduced in Section~\ref{sec:BNNs}.
\subsection{Binarized Neural Networks}\label{sec:BNNs}
A BNN \cite{hubara2016binarized} is a feedforward network where weights and activations are mainly binary. \cite{narodytska2018verifying} describes BNNs as sequential composition of blocks, each block consisting of linear and non-linear transformations. One could distinguish between \emph{internal} and \emph{output blocks}. 

There are typically several \emph{internal blocks}. The layers of the blocks are chosen in such a way that the resulting architecture fulfills the requirements of accuracy, model size, number of parameters, for example. Typical layers in an internal block are:
\begin{inparaenum}[\itshape 1)\upshape]
\item linear transformation (\textsc{LIN})
\item binarization (\textsc{BIN})
\item max pooling (\textsc{MP})
\item batch normalization (\textsc{BN}).
\end{inparaenum}
A linear transformation of the input vector can be based on a fully connected layer or a convolutional layer. In our case is a convolution layer since our experiments have shown that a fully connected layer can not synthesize well the features of traffic signs, therefore, the accuracy is low. The linear transformation is followed either by a binarization or a max pooling operation. Max pooling helps in reducing the number of parameters. One can swap binarization with max pooling, the result would be the same. We use this sequence as Larq~\cite{geiger2020larq}, the library we used in our experiments, implements convolution and binarization in the same function. Finally, scaling is performed with a batch normalization operation~\cite{ioffe2015batch}. 

There is \emph{one output block} which produces the predictions for a given image. It consists of a dense layer that maps its input to a vector of integers, one for each output label class. It is followed by function which outputs the index of the largest entry in this vector as the predicted label.

We make the observation that, if the MP and BN layers are omitted, then the input and output of the internal blocks are binary, in which case, also the input to the output block. The input of the first block is never binarized as it drops down drastically the accuracy.
\subsection{Properties of (Binarized) Neural Networks: Robustness}\label{sec:ThVNNs}
\emph{Robustness} is a fundamental property of neural networks that refers to their ability to maintain stable and accurate outputs in the presence of perturbations or adversarial inputs. Adversarial inputs are intentionally crafted inputs designed to deceive or mislead the network's predictions. 

As defined by~\cite{narodytska2018verifying}, \emph{local robustness} ensures that for a given input $x$ from a set $\chi$, the neural network $F$ remains unchanged within a specified perturbation radius $\epsilon$, implying that small variations in the input space do not result in different outputs. The output for the input $x$ is represented by its label $l_x$. We consider $L_\infty$ norm defined as $||x||_{\infty}=\sup\limits_{n}|x_{n}|$, but also other norms can be used, e.g.~$L_0$~\cite{ruan2018global}.
\begin{defn}[Local robustness.] \label{def:localRobustness}
A feedforward neural network $F$ is locally $\epsilon$-robust for an input $x, x \in \chi$, if there does not exist $\tau, ||\tau||_\infty \leq \epsilon$, such that $F(x + \tau) \neq l_x$.
\end{defn}

\emph{Global robustness}~\cite{narodytska2018verifying} is an extension of the local robustness and it is defined as the expected maximum safe radius over a given test dataset, representing a collection of inputs. 
\begin{defn}[Global robustness.] A feed-forward neural network $F$ is globally $\epsilon$-robust if for any $x, x \in \chi$, and $\tau, ||\tau||_\infty \leq \epsilon$, we have that $F(x + \tau) = l_x$.
\end{defn}

The definitions above can not be used in a computational setting. Hence, ~\cite{katz2022reluplex} proposes Definition~\ref{def:localRobustnessComputational}  for local robustness which is equivalent to Definition ~\ref{def:localRobustness}.
\begin{defn}[Local robustness.] \label{def:localRobustnessComputational}
A network is $\epsilon$-locally robust in the input $x$ if for every $x'$, such that $||x - x'||_\infty \leq \epsilon$, the network assigns the same label to $x$ and $x'$.
\end{defn}
For our setting, the input is an image represented as a vector with values represented by the pixels. Hence, the inputs are the vector $x$ and the perturbation $\epsilon$.

This formula can also be applied to all inputs simultaneously (all images from test set of the dataset), in that case \emph{global robustness} is addressed. However, the number of parameters involved in checking \emph{global robustness} property increases enormously. Hence, in this paper, the benchmarks propose verification of local robustness only.
\section{Anatomy of the Binarized Neural Networks}\label{sec:anatomy}
\begin{figure}[h]
  \centering
    \includegraphics[width=0.7\textwidth]{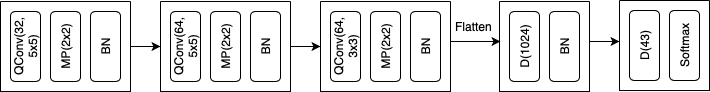}
    \caption{Accuracy Efficient Architecture for GTSRB and Belgium dataset}
    \label{fig:Acc-Efficient-Arch-GTSRB-Belgium}
\end{figure}

\begin{figure}[h]
  \centering
    \includegraphics[width=0.7\textwidth]{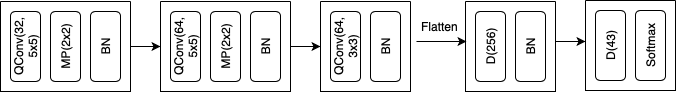}
    \caption{Accuracy Efficient Architecture for Chinese dataset}
    \label{fig:Acc-Efficient-Arch-Chinese}
\end{figure}

For benchmarking, we propose the two BNNs architectures for which we obtained the best accuracy~\cite{postovan2023architecturing}, as well an additional one. More precisely, the best accuracy for GTSRB and Belgium datasets is $96,45\%$ and $88,17\%$, respectively, and was obtained for the architecture from Figure~\ref{fig:Acc-Efficient-Arch-GTSRB-Belgium}, with input size 64$\times$64 (see Table~\ref{tab:QConv_32_5_MP_2_BN_QConv_64_5_MP_2_BN_QConv_64_3_MP_2_BN_(Dense_???)*_Dense_43}). The number of parameters is almost $2$M and the model size $225,67$~KiB (for the binary model) compared to $6932,48$ KiB (for the Float-32 equivalent). 
\begin{table}[h]
\centering
\scriptsize
\caption{Best results for the architecture from Figure~\ref{fig:Acc-Efficient-Arch-GTSRB-Belgium}. Dataset for train: GTSRB.}
\label{tab:QConv_32_5_MP_2_BN_QConv_64_5_MP_2_BN_QConv_64_3_MP_2_BN_(Dense_???)*_Dense_43}
\begin{tabular}{|l|c|ccc|ccc|cc|}
\hline
\multirow{2}{*}{\textbf{Input size}} &
  \multicolumn{1}{c|}{\multirow{2}{*}{\textbf{\#Neur}}} &
  \multicolumn{3}{c|}{\textbf{Accuracy}} &
  \multicolumn{3}{c|}{\textbf{\#Params}} &
  \multicolumn{2}{c|}{\textbf{Model Size (in KiB)}} \\ \cline{3-10} 
 &
  \multicolumn{1}{c|}{} &
  \multicolumn{1}{c|}{\textbf{German}} &
  \multicolumn{1}{c|}{\textbf{China}} &
  \multicolumn{1}{c|}{\textbf{Belgium}} &
  \multicolumn{1}{c|}{\textbf{Binary}} &
  \multicolumn{1}{c|}{\textbf{Real}} &
  \multicolumn{1}{c|}{\textbf{Total}} &
  \multicolumn{1}{c|}{\textbf{Binary}} &
  \multicolumn{1}{c|}{\textbf{Float-32}} \\ \hline
64px $\times$ 64px &
  1024 &
  \multicolumn{1}{c|}{\textbf{96.45}} &
  \multicolumn{1}{c|}{\textbf{81.50}} &
  \textbf{88.17} &
  \multicolumn{1}{c|}{1772896} &
  \multicolumn{1}{c|}{2368} &
  1775264 &
  \multicolumn{1}{c|}{225.67} &
  6932.48 \\ \hline
\end{tabular}
\end{table}
The best accuracy for Chinese dataset ($83,9\%$) is obtained by another architecture, namely from Figure~\ref{fig:Acc-Efficient-Arch-Chinese}, with input size 48$\times$48 (see Table~\ref{tab:QConv_32_5_MP_2_BN_QConv_64_5_MP_2_BN_QConv_64_3_BN_(Dense_???)*_Dense_43}). This architecture is more efficient from the point of view of computationally limited devices and formal verification having $900$k parameters and $113,64$ KiB (for the binary model) and $3532,8$ KiB (for the Float-32 equivalent). Also, the second architecture gave the best average accuracy and the decrease in accuracy for GTSRB and Belgium is small, namely $1,17\%$ and $0,39\%$, respectively.
\begin{table}[h]
\centering
\scriptsize
\caption{Best results for the architecture from Figure~\ref{fig:Acc-Efficient-Arch-Chinese}. Dataset for train: GTSRB.}
\label{tab:QConv_32_5_MP_2_BN_QConv_64_5_MP_2_BN_QConv_64_3_BN_(Dense_???)*_Dense_43}
\begin{tabular}{|l|c|ccc|ccc|cc|}
\hline
\multirow{2}{*}{\textbf{Input size}} &
  \multicolumn{1}{c|}{\multirow{2}{*}{\textbf{\#Neur}}} &
  \multicolumn{3}{c|}{\textbf{Accuracy}} &
  \multicolumn{3}{c|}{\textbf{\#Params}} &
  \multicolumn{2}{c|}{\textbf{Model Size (in KiB)}} \\ \cline{3-10} 
 &
  \multicolumn{1}{c|}{} &
  \multicolumn{1}{c|}{\textbf{German}} &
  \multicolumn{1}{c|}{\textbf{China}} &
  \multicolumn{1}{c|}{\textbf{Belgium}} &
  \multicolumn{1}{c|}{\textbf{Binary}} &
  \multicolumn{1}{c|}{\textbf{Real}} &
  \multicolumn{1}{c|}{\textbf{Total}} &
  \multicolumn{1}{c|}{\textbf{Binary}} &
  \multicolumn{1}{c|}{\textbf{Float-32}} \\ \hline
48px $\times$ 48px &
  256 &
  \multicolumn{1}{c|}{\textbf{95.28}} &
  \multicolumn{1}{c|}{\textbf{83.90}} &
  \textbf{87.78} &
  \multicolumn{1}{c|}{904288} &
  \multicolumn{1}{c|}{832} &
  905120 &
  \multicolumn{1}{c|}{113.64} &
  3532.80 \\ \cline{2-10}  \hline
\end{tabular}
\end{table}

\begin{figure}[h]
  \centering
    \includegraphics[width=0.3\textwidth]{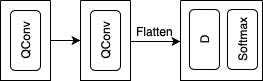}
    \caption{XNOR(QConv) architecture}
    \label{fig:XNOR(QConv)-arch}
\end{figure}

One could observe that the best architectures were obtained for input size images 48x48 and 64x64 pixels with max pooling and batch normalization layers which reduce the number of neurons, namely perform scaling which leads to good accuracy. We also propose for benchmarking an XNOR architecture, i.e. containing only binary parameters, (Figure \ref{fig:XNOR(QConv)-arch}) for which we obtained the best results for input size images 30x30 pixels and GTSRB (see Table \ref{tab:XNOR(QConv)}).
\begin{table}[h]
\caption{\textsc{XNOR(QConv)} architecture. Image size: 30px $\times$ 30px. Dataset for train and test: GTSRB.}
\label{tab:XNOR(QConv)}
\centering
\scriptsize
\begin{tabular}{|lcccc|}
\hline
\multicolumn{1}{|l|}{\multirow{2}{*}{\textbf{Model description}}}                                            & \multicolumn{1}{c|}{\multirow{2}{*}{\textbf{Acc}}} & \multicolumn{1}{c|}{\multirow{2}{*}{\textbf{\begin{tabular}[c]{@{}c@{}}\#Binary \\ Params\end{tabular}}}} & \multicolumn{2}{c|}{\textbf{Model Size (in KiB)}}        \\ \cline{4-5} 
\multicolumn{1}{|l|}{}                                                                               & \multicolumn{1}{c|}{}                              & \multicolumn{1}{c|}{}                                                      & \multicolumn{1}{c|}{\textbf{Binary}} & \textbf{Float-32} \\ \hline
\multicolumn{1}{|l|}{\begin{tabular}[c]{@{}l@{}}QConv(16, 3$\times$3), QConv(32, 2$\times$2), D(43)\end{tabular}}          & \multicolumn{1}{c|}{81.54}                         & \multicolumn{1}{c|}{1005584}                                                                              & \multicolumn{1}{c|}{122.75}          & 3932.16           \\ \hline
\end{tabular}
\end{table}
\section{Model and Property Specification: \textsc{VNN-LIB} and ONNX Formats}\label{sec:VNNLibFormat}
\noindent The \textsc{VNN-LIB} (Verified Neural Network Library) format \cite{guidotti023years} is a widely used representation for encoding and exchanging information related to the verification of neural networks. It serves as a standardized format that facilitates the communication and interoperability of different tools and frameworks employed in the verification of neural networks.

The \textsc{VNN-LIB} format typically consists of two files that provide a detailed specification of the neural network model (see Section~\ref{sec:modelRepresentation}), along with relevant properties and constraints (see Section~\ref{sec:propertySpec}). These files encapsulate important information, including the network architecture, weights and biases, input and output ranges, and properties to be verified.
\subsection{Model Representation}\label{sec:modelRepresentation}
In machine learning, the representation of models plays a vital role in facilitating their deployment and interoperability across various frameworks and platforms. One commonly used format is the \textsc{H5} format, which is an abbreviation for \emph{Hierarchical Data Format version 5}. The \textsc{H5} format provides a structured and efficient means of storing and organizing large amounts of data, including the parameters and architecture of machine learning models. It is widely supported by popular deep learning frameworks, such as TensorFlow and Keras, allowing models to be saved, loaded, and shared in a standardized manner.

However, while the \textsc{H5} format serves as a convenient model representation for specific frameworks, it may lack compatibility when transferring models between different frameworks or performing model verification. This is where the \emph{Open Neural Network Exchange} (\textsc{ONNX}) format comes into play. \textsc{ONNX} offers a vendor-neutral, open-source alternative that allows models to be represented in a standardized format, enabling seamless exchange and collaboration across multiple deep learning frameworks.

The VNN-LIB format, which is used for the formal verification of neural network models, leverages \textsc{ONNX} as its underlying model representation.
\subsection{Property specification}\label{sec:propertySpec}
For property specification, VNN-LIB standard uses the SMT-LIB format. The \textsc{SMT-LIB} (Satisfiability Modulo Theories-LIBrary) language~\cite{barrett2010smt} is a widely recognized formal language utilized for the formalization of Satisfiability Modulo Theories (\textsc{SMT}) problems. 

A VNN-LIB file is structured as follows\footnote{See, e.g. \url{https://github.com/apostovan21/vnncomp2023/blob/master/vnnlib/model_30_idx_1678_eps_1.00000.vnnlib}} and the elements involved have the following semantics for the considered image classification task:
\begin{enumerate}
\itemsep0em
    \item definition of input variables representing the values of the pixels $X_i$ ($i = \overline{1,P}$, where $P$ is the dimension of the input image: $N \times M \times 3$ pixels). For the file above, there are $2700$ variables as the image has dimension $30 \times 30$ and the channel used is RGB.
    \item definition of the output variables representing the values $Y_j$ ($j = \overline{1,L}$, where $L$ is the number of classes of the images in the dataset). For the file above, there are $43$ variables as the GTSRB categorises the traffic signs images into $43$ classes.
    \item bounding constraints for the variables input variables. Definition \ref{sec:modelRepresentation} is used for generating the property taking into account that vector $x$ (its elements are the values of the pixels of the image) and $\varepsilon$ (perturbation) are known. For example, if $\varepsilon=10$ and the value of the pixel $X_{2699}'$ of the image with index $1678$ from GTSRB is $24$, the generated constraints for finding the values of the perturbed by $\varepsilon$ pixel $X_{2699}$ for which the predicted label still holds is:
    \begin{verbatim}
    (assert (<= X_2699 34.00000000))
    (assert (>= X_2699 14.00000000))
    \end{verbatim}
    \item constraints involving the output variables assessing the value of the output label. For example, if the verification problem is formulated as: \emph{Given the image with index $1678$, the perturbation $\varepsilon~=~10$ and the trained model, find if the perturbed images are in class $38$}, the generated constraints are as follows which actually represents the negation of the property to be checked:
    \begin{verbatim}
    (assert (or (>= Y_0 Y_38)
                ...
                (>= Y_37 Y_38)
                (>= Y_39 Y_38)
                ...
                (>= Y_42 Y_38)))
    \end{verbatim}
\end{enumerate}
\section{Benchmarks Proposal and Experimental Results of the VNN-COMP 2023}\label{sec:Bench&ExpResults}
To meet the requirements of the VNN-COMP 2023, the benchmark datasets must conform to the ONNX format for defining the neural networks, while the problem specifications are expected to adhere to the VNN-LIB format. Therefore, we have prepared a benchmark set comprising the BNNs introduced in Section~\ref{sec:anatomy} that have been encoded in the ONNX format. 
In order to assess the adversarial robustness of these networks, the problem specifications encompassed perturbations within the infinity norm around zero, with radius denoted as $\epsilon = \{1, 3, 5, 10, 15\}$. To achieve this, we randomly selected three distinct images from the test set of the GTSRB dataset for each model and have generated the \textsc{VNNLIB} files for each epsilon in the set, in the way we ended up having 45 \textsc{VNNLIB} files in total. We were constrained to generate the small benchmark which includes just 45 \textsc{VNNLIB} files because of the total timeout which should not exceed 6 hour, this is the maximum timeout for a solver to address all instances, consequently a timeout of 480 seconds was allocated for each instance. For checking the generated VNNLIB specification files for submitted in the VNNCOMP 2023 as specified above as well as to generate new ones you can check \url{https://github.com/apostovan21/vnncomp2023}.

Our benchmark was used for scoring the competing tools. The results for our benchmark, as presented by the VNN-COMP 2023 organizers, are presented in Table~\ref{tab:res-Traffic-Signs-Recognition-benchmark}.
\begin{table}[h]
\centering
\scriptsize
\caption{VNN-COMP 2023 Results for Traffic Signs Recognition Benchmark}
\label{tab:res-Traffic-Signs-Recognition-benchmark}
\begin{tabular}{cccccccc}
\textbf{\#} & \textbf{Tool}    & \textbf{Verified} & \textbf{Falsified} & \textbf{Fastest} & \textbf{Penalty} & \textbf{Score} & \textbf{Percent} \\ \hline
1           & Marabou          & 0                 & 18                 & 0                & 1                & 30             & 100\%            \\
2           & PyRAT            & 0                 & 7                  & 0                & 1                & -80            & 0\%              \\
3           & NeuralSAT        & 0                 & 31                 & 0                & 4                & -290           & 0\%              \\
4           & alpha-beta-CROWN & 0                 & 39                 & 0                & 3                & -60            & 0\%             
\end{tabular}
\end{table}

The meaning of the columns is as follows. Verified is number of instances that were UNSAT (no counterexample) and proven by the tool. Falsifieid is number that were SAT (counterexample was found) and reported by the tool. Fastest is the number where the tool was fastest (this did not impact the scoring in this year competition). Penalty is the number where the tool gave the incorrect result or did not produce a valid counterexample. Score is the sum of scores ($10$ points for each correct answer and $-150$ for incorrect ones). Percent is the score of the tool divided by the best score for the benchmark (so the tool with the highest score for each benchmark gets $100$) and was used to determine final scores across all benchmarks.

Currently, we are investigating if the number of solved instances could be higher if the time is increased (the deadline used was 8 minutes). Also, it is interesting why the tools gave incorrect results for some benchmarks.
\section{Conclusions}
Building upon our prior study that introduced precise binarized neural network models for traffic sign recognition, this study presents standardized challenges to gauge the resilience of these networks to local variations. These challenges were entered into the VNN-COMP 2023 evaluation, where 4 out of 7 tools produced results. Our current emphasis is on investigating the potential for solving more instances by extending the time limit (formerly set at 8 minutes). Additionally, we are keen to comprehend the factors contributing to incorrect outputs from the tools on specific benchmark tasks.
\section*{Acknowledgements}
\thanks{This work was supported by a grant of the Romanian National Authority for Scientific Research and Innovation, CNCS/CCCDI - UEFISCDI, project number PN-III-P1-1.1-TE-2021-0676, within PNCDI~III.}
\bibliographystyle{eptcs}
\bibliography{biblio}
\end{document}